# Empathetic Conversational Systems: A Review of Current Advances, Gaps, and Opportunities

Aravind Sesagiri Raamkumar and Yinping Yang

**Abstract**— Empathy is a vital factor that contributes to mutual understanding, and joint problem-solving. In recent years, a growing number of studies have recognized the benefits of empathy and started to incorporate empathy in conversational systems. We refer to this topic as empathetic conversational systems. To identify the critical gaps and future opportunities in this topic, this paper examines this rapidly growing field using five review dimensions: (i) conceptual empathy models and frameworks, (ii) adopted empathy-related concepts, (iii) datasets and algorithmic techniques developed, (iv) evaluation strategies, and (v) state-of-the-art approaches. The findings show that most studies have centered on the use of the EMPATHETICDIALOGUES dataset, and the text-based modality dominates research in this field. Studies mainly focused on extracting features from the messages of the users and the conversational systems, with minimal emphasis on user modeling and profiling. Notably, studies that have incorporated emotion causes, external knowledge, and affect matching in the response generation models, have obtained significantly better results. For implementation in diverse real-world settings, we recommend that future studies should address key gaps in areas of detecting and authenticating emotions at the entity level, handling multimodal inputs, displaying more nuanced empathetic behaviors, and encompassing additional dialogue system features.

✦

## 1 INTRODUCTION

Conversational artificial intelligence (CAI) has become a lucrative area for research and commercial applications in the form of personalized digital assistants, virtual assistants, cobots, and chatbots [1]. They have proliferated several application domains ranging from daily life, commerce, business support, education, to healthcare [2], [3]. CAI research spans multiple topics encompassing conversational chatbot and dialogue systems [4], [5], conversational recommender systems [6], conversational search systems [7], and conversational question and answering systems [8].

Although each of these topics has specific foci, a full-scale CAI implementation would benefit from an effective integration of research ideas and outputs from these topics as the user's requirements in the real-world transition between multiple states. For example, when a user interacts with a CAI-based customer service agent, the user expresses the request to the agent in natural language (applicable to conversational search), followed by further questions and clarifications from the agent (applicable to chit-chat dialogue systems) so that the agent fully understands the request. After completing the inquiries, the agent has all the details to convert the user's request to a query representation which is used to retrieve the relevant data from the databases. Subsequently, the agent might provide either a factual response (applicable to conversational question and answering ) or provide recommendations (applicable to conversational recommender systems) based on the request criteria. Hence, we opine that a complete CAI implementation should have capabilities from all the related research topics to simulate an effective real-world experience. As humans are a vital part of the loop in CAI studies, behavioral research on human-chatbot interactions [3], [9] and information-seeking strategies [10] have likewise been conducted.

A key objective in CAI research is to humanize systems to facilitate better and more meaningful engagement with humans [11]. Researchers have since developed emotionally-aware systems to detect sentiments and emotions from human expressions and generate emotional responses [12]. The implementations rely on sentiment analysis [13] and emotion recognition [14] algorithms to identify user messages' prevailing sentiments and emotions. Although identifying sentiments and emotions is a constructive step towards building an effective conversation, engaging humans with empathetic responses has proven more successful in CAI studies from domains such as health [15] and marketing [16]. These studies highlight the positive experience of participants when they perceived affective empathy from the agents. The empathetic conversational feature was found to contribute to bridging the human-AI gap.

Based on an analysis of about 43 definitions, Cuff et al. [17] summarized empathy as "*an emotional response (affective), dependent upon the interaction between trait capacities and state influences. Empathic processes are automatically elicited but are also shaped by top-down control processes. The resulting emotion is similar to one's perception (directly experienced or imagined) and understanding (cognitive empathy) of the stimulus emotion, with the recognition that the source of the*

---

• *A. Sesagiri Raamkumar and Y. Yang are with the Institute of High Performance Computing, Agency for Science, Technology and Research (A\*STAR), #16-16 Connexis North, 1 Fusionopolis Way, Singapore 138632 E-mail: aravindsr@ihpc.a-star.edu.sg; yangyp@ihpc.a-star.edu.sg.*

*emotion is not one's own"*. Empathy is considered a necessary behavior, and studies have been conducted to improve empathy amongst humans in different settings [18], [19]. There are also different types of empathy, namely affective empathy, cognitive empathy, and compassionate empathy [20]. While affective empathy and cognitive empathy are about mirroring and understanding others' feelings, respectively, compassionate empathy is about providing socially desirable responses to others' feelings.

Computational modeling of empathy aids in better understanding human relations [21]. Computational and theoretical empathy modeling studies have been conducted with variations in three main components – emotional communication competence, emotion regulation, and cognitive mechanisms [22]. Empathetic behaviors differ based on the mechanisms related to these three components. The cataloged behaviors are mirroring, affective matching, empathic concern, consolation, altruistic helping, and perspective-taking [22]. An ideal empathetic CAI system is expected to exhibit these behaviors depending on the conversation scenario.

The recent advancements in deep learning and natural language processing (NLP) have accelerated the research in CAI with new trends in multimodal, multitask, and long-term goal handling systems [23]. Correspondingly, there has been increased interest in empathetic response generation approaches since the introduction of the EMPATHETICDIALOGUES dataset, and corresponding response generation models [24]. CAI systems can be trained to show empathy towards human feelings in text-based conversations. Subsequently, empathetic response generating CAI systems research expanded with new datasets [25]–[28] and enhanced response generation models [29], [30]. CAI dialog systems can be classified into three types – (i) task-oriented, (ii) conversational, and (iii) interactive questions and answering [31]. Based on the existing empathetic response generating systems studies covered in this review, the research topic can be characterized as conversational systems since the dialogue structure is unstructured, the number of turns is multiple, the length of the dialogues is long, and also because there is no specific task being completed by the CAI systems.

In this paper, we focus on the notion of empathetic conversational systems (ECS) as a class of CAI systems which seek to incorporate empathy. This helps to differentiate ECS from the term embodied conversational agents, or ECA [32].

## 1.1 Need for this Review

Previous ECS studies have shown much promise in conceptualizing frameworks, preparing datasets, training models, and designing algorithms for embedding empathy in CAI systems. In one of the earliest surveys in this field by Paiva et al. published in 2017 [33], computational empathy simulation and triggering mechanisms in virtual agents and robots were surveyed. During this survey's publication period, systems primarily employed rule-based and heuristics-based approaches instead of the current deep learning-based natural language generation

(NLG) approaches. In Spring et al. [34], ECS studies are reviewed using a framework that comprises four stages, namely emotion expression, emotion detection classification, response generation, and response expression. ECS studies are surveyed from the perspective of functions by Ma et al. [35], and three types of dialogue systems are surveyed in the purview of empathetic dialogue systems. These dialogue system types are affective dialogue systems, personalized dialogue systems, and knowledge-based dialogue systems. In Wardhana et al. [36], the empathetic dialogue characteristics, dialogue system models, and statistical inference techniques have been reviewed.

Despite these valuable reviews, there has been a distinct lack of systematic insights which have investigated empathy incorporation in ECS models in an in-depth manner. In particular, the existing reviews have not included examination about the conceptual empathy models and empathy-related concepts that have been operationalized in empirical ECS studies. Moreover, the prominent datasets used in the studies have also not been covered. Empathy is a multidimensional concept, and multiple peripheral sub-concepts are at play in human-agent interaction. A critical review of ECS studies is warranted from this frame of reference.

## 1.2 Objectives of the Review

Our objective is to critically review existing ECS studies, and the conceptual frameworks used to understand how different studies have attempted to invoke empathy in systems. This will help delineate the advancements in this research topic so that the gaps and opportunity areas can be identified. Specifically, we examine the following questions.

1) What conceptual empathetic models are used to guide the design and development of the existing ECS implementations? What are the different empathy and empathy-related concepts operationalized by the current ECS studies?

2) Which datasets are developed and used in ECS studies, and how are these datasets generated?

3) What and how algorithmic models are used in ECS studies for request processing and response generation activities?

4) What are the evaluation approaches and metrics used in ECS studies?

5) What are the most notable SOTA approaches in existing ECS?

This paper focuses on studies in which the systems have been trained to exhibit empathy explicitly. The studies which focus on providing emotional responses to users' messages (e.g., [37]) have been reviewed in earlier reviews [12], [34], and are not covered in this review.

## 2 METHOD

### 2.1 Article Selection

Multiple digital literature searches were conducted across Scopus, Google Scholar, and IEEE Xplore digital library.

The main search strategy was the discovery of papers relevant to the topic using the following query phrases without any publication year filter applied: *"empathetic conversational"*, *"empathetic chatbot"*, *"empathetic dialogue"*, *"empathetic dialog"*, *"empathic conversational"*, *"empathic chatbot"*, *"empathic dialogue"*, *"empathic dialog"*, *"empathy conversation"*, *"empathy chatbot"*, *"empathy dialogue"*, *"empathy dialog"*, *"emotion conversation"*, *"emotion chatbot"*, *"emotion dialogue"*, *"emotion dialog"*, *"empath\* AI"*, *"empath\* agents"*, *"empath\* artificial intelligence"*. Citations and references trail were performed on the initially identified papers using the papers' citations and references as a starting point and a total of 112 papers were identified. These papers were next assessed for relevance through analysis of article title and abstract fields. As a result, 66 papers were not considered relevant to the review. The full text of the remaining papers was scanned.

In total, we first considered 46 papers for our in-depth review. It is useful to note that we subsequently found few papers that were published as a part of the Alexa SocialBot challenge [38]. Although these papers [39], [40] have proposed models that were trained on ECS datasets such as the EMPATHETICDIALOGUES dataset [24] along with other datasets, the algorithms were not specifically conceptualized for empathetic response generation. Hence, these two papers were removed. Figure 1 depicts the article selection process flow.

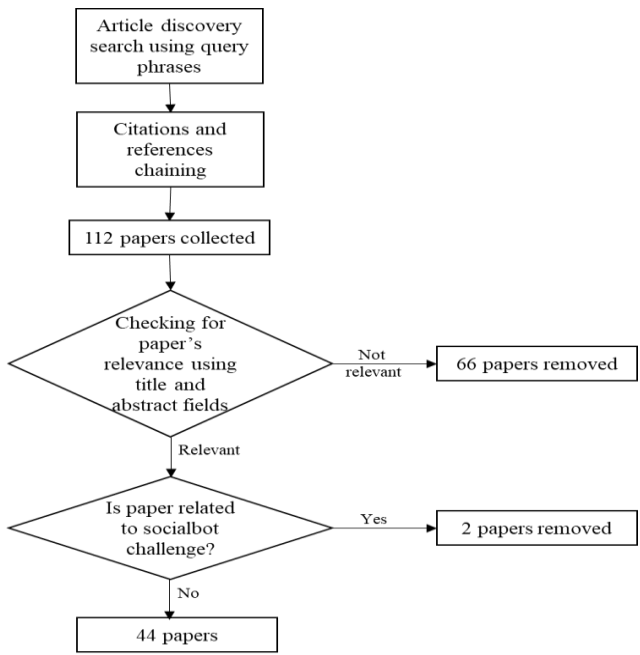

Fig. 1: Article Selection Process Flow

## 2.2 Data Abstraction

The final list of selected papers was then subjected to a data abstraction exercise. The features used in the data abstraction include dataset name, dataset source algorithms (for emotion/sentiment recognition and empathetic response generation), study objectives, related empathy concept(s), evaluation flag, offline evaluation metrics, user evaluation metrics, ablation study flag apart from the primary paper metadata fields such as publication venue, article type and publication year. The list of ECS empirical studies along with the extracted features has been included in Appendix A.

## 3 CONCEPTUAL EMPATHY MODELS

Empathy, the capacity to relate to others, has been credited as a vital factor in improved relationships and outcomes based on research in several disciplines, including industry and organizational psychology, leadership development, social psychology, negotiations, neuroscience, and mental health [41]–[43]. It is a complex, multidimensional, and high-order social intelligence skill. According to Goleman [44], [45], empathy involves at least three facets: cognitive empathy, affective empathy, and empathic concern.

*Cognitive empathy* is the ability to understand another person's point of view or perspective. Cognitive empathy is closely related and used interchangeably with perspective-taking. It is about "putting oneself in others' shoes".

*Affective empathy*, or emotional empathy, is the ability to feel what someone else feels. Affective or emotional empathy is about "your pain in my heart". In complex situations, feeling fast without thinking deeply is an essential skill that is linked to human evolution.

*Compassionate empathy*, or empathic concern, is the ability to sense what another person needs from you and do something helpful. This facet of empathy goes beyond perspective-taking and sharing others' feelings but demonstrates helpful behaviors that incorporate the information about others for more effective problem-solving.

Computational approaches to incorporating empathy have presented a variety of empathy conceptualizations. De Waal's three-layer Russian-doll empathy model [46] forms the basis for the subsequent computational empathy models and frameworks. According to this model, empathy is perceived as a shared emotional experience between two persons when one person happens to feel a similar emotion as the other person. One person's representations of the emotional state are spontaneously activated when this person pays attention to the emotional state of the other person. At the lowest layer of this model, the model places an affective matching component exhibited by mimicry. The middle layer is for consolation, which is exhibited by sympathetic concern. The upper layer is represented by targeted helping, which is exhibited by perspective-taking.

Yalcin et al. [32] put forth a framework for equipping ECS with real-time multimodal empathic interaction capabilities based on their model of empathy. The model comprises a three-level hierarchy encompassing communication competence, affect regulation, and cognitive mechanism [47]. The empathy framework [32] includes perceptual, behavior controller, and behavior manager modules. The perceptual module collects the user's input through both audio-video signals. It then sends these signals to the emotion recognition sub-module for ascertaining the emotions in the user input. The processed data is subsequently passed to the behavior controller, where the user intent is thoroughly analyzed. The empathy mechanism sub-module in the module also functions at three levels similar to the empathy model [47] – low level, mid-level and high-level empathetic behavior. At the lowest level, mimicry and affect matching is exhibited while emotion regulation is enabled at the mid-level by considering user mood, personality, and likes/dislikes preferences. Cognitive processes are at the highest level of empathy where the user goals and context are considered. The empathetic response is embedded with the factual response, and it is sent back to the users through the behavior manager module in the framework.

Ab Aziz et al. [48] proposed another conceptual design model for incorporating empathy in CAI agents. The model comprises five main modules, namely i) sensing, ii) emotion analysis, personality, and event evaluation, iii) empathy analytics and behavior selection, iv) stress analytics and support, and v) feedback. The sensing module receives the user input through voice, visuals, and touch-points facilitated by a touch screen interface. The input is passed over to the emotion analysis module for face recognition, feature extraction, and emotion recognition. The input is also processed by the personality and event evaluation on a parallel front. The output of these modules is sent to the empathy analytics module. This module is based on another proposed integrated empathy model, which combines the Belief Desire Intention (BDI) model [49], three types of empathy (i.e., emotional, cognitive, and compassionate), and the theory of mind [50]. The next module is the behavior section module which comprises a database of corresponding behaviors and actions. The behaviors and actions are augmented with cues from the stress analytic and support module. The final output is relayed to the users through the feedback module in voice and screen output.

## 4   EMPATHETIC CONVERSATIONAL SYSTEMS (ECS)

The majority of the existing empirical ECS studies involve supervised deep learning techniques; hence the training dataset and the architecture are the essential aspects. The studies are summarized in this section based on the following characteristics: empathy-related concepts, algorithms/techniques in the architecture, datasets, and evaluation strategies. We have not considered studies that employ commercial off-the-shelf tools for developing CAI systems, as their underlying techniques are not in the public domain.

### 4.1 Empathy Related Concepts Representation

At a fundamental level, understanding human emotions and subsequently providing empathetic responses are the main functions of an ECS. Studies can be synthesized using these two functions. The functions need to be more detailed to facilitate a better representation of the different empathy components in ECS studies. Yalcin et al. [21] proposed three empathy components as part of an empathy model: communication competence, emotion regulation, and cognitive mechanisms to study empathic behavior. Spring et al. [34], on the other hand, proposed a four-stage model to study empathic chatbots. The stages are emotion expression, emotion detection, response generation, and response expression.

While the Yalcin et al. [21] model is suitable for studying empathic behavior, the Spring et al. [34] model could be considered a pragmatic model for categorizing ECS studies during that earlier period when research in ECS was still at a nascent stage. We are using a component-based model for categorizing the existing ECS studies for this review paper. The components are the two actors (user, ECS) and their messages (including requests and responses). These components cover all the empathy-related concepts represented in the existing ECS studies. We have added one peripheral component, namely Knowledge Bases, in this model since it has been utilized in a few studies. In Figure 2, we have illustrated the concepts from the studies and mapped them to the actors or the components of the message based on their relation.

### User and User Messages

User-based activities such as user modeling and user-based filtering have not been employed in ECS studies thus far. Hence, very few concepts pertain to the user actor in Fig. 2. The majority of the concepts are on the user messages component ($n = 32$). The list of concepts is included in Table 1.

*(a) Emotion*: Detecting emotion class in user messages has been a de-facto step towards empathetic response generation, mainly because of the usage of the EMPATHETICDIALOGUES dataset [24]. Emotion class is detected in 23 studies. As per the seminal study [24], emotion class identification facilitates the selection of appropriate empathetic responses. These ECS studies considered a total of 32 emotion classes. Therefore, most emotional situations were covered. However, no ECS study has attempted to validate the sufficiency or redundancy of the 32 emotion classes. In addition, none of the ECS studies have leveraged algorithms or models that venture beyond emotion classification, such as affect or emotion intensity quantification [51].

*(b) Sentiment*: Understanding emotions in user messages provides more insights than sentiments since sentiments, focusing on positive, negative, and neutral classification, are not as descriptive and actionable as emotions [52]. However, a few studies have identified sentiments in user messages to respond empathetically [53]–[55]. With the current advancements in emotion classification research studies and the availability of pre-trained emotion classification models [56], [57], the option of using a sentiment classifier in ECS architecture design, might not bring forth the best results in future ECS studies.

*(c) Positive and Negative Emotion Clusters*: After identifying emotion classes, one ECS study attempted clustering positive and negative emotions separately to generate empathetic responses [58]. As per the study findings, this emotion grouping/clustering approach is coupled with the emotion mimicry approach to produce better evaluation results, albeit the resultant model is benchmarked against only one ECS study [59].

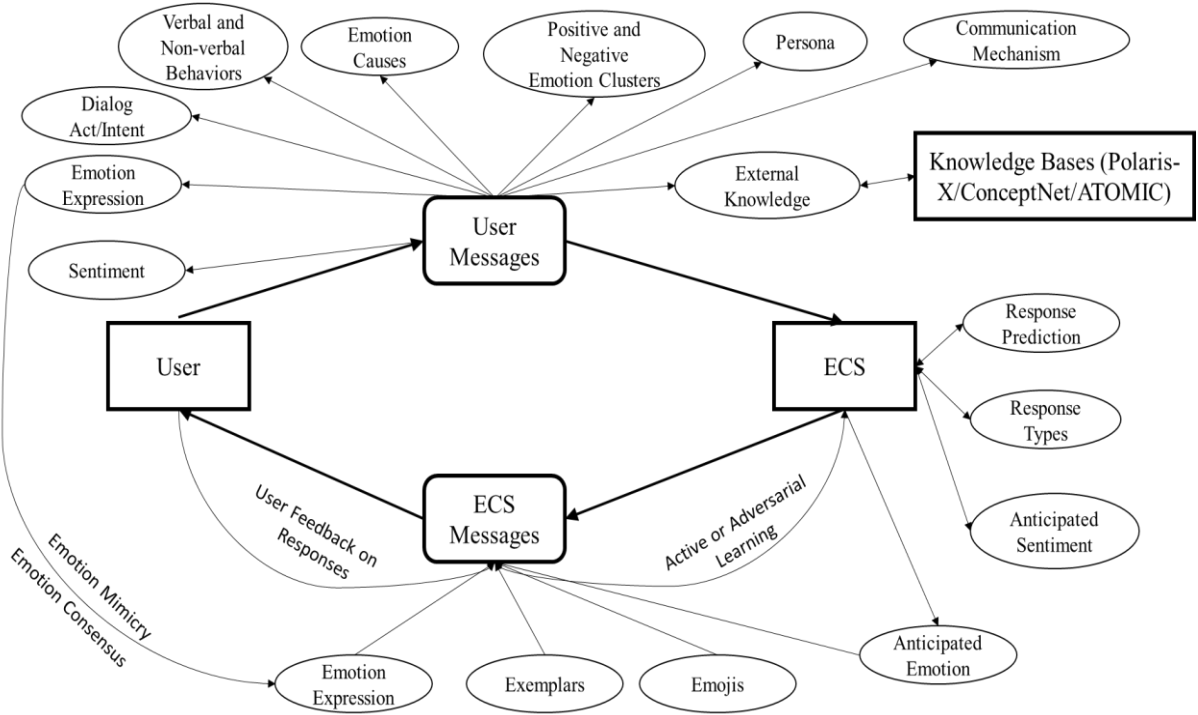

Fig 2: Operationalized Empathy-related Concepts in ECS Studies

TABLE 1
EMPATHY-RELATED CONCEPTS FROM ECS STUDIES

| Concept | Definition | Improvements provided to ECS Studies | Related ECS Studies |
|---------|------------|--------------------------------------|---------------------|
| User Messages | | | |
| Emotion | Reaction (e.g., joy or fear) experienced as a feeling directed toward an event, a person/agent, or an object | Understanding the user's emotion is a necessary step before exhibiting empathy | [24], [29], [30], [52], [55]–[58], [65], [66], [73], [75], [76], [81], [82], |
| Sentiment | Mental attitude (e.g., positive, negative, neutral) produced by feelings and perception | Helps in understanding the user's attitude so that affective matching mechanisms can be activated | [53]–[55] |
| Positive and Negative Emotion Clusters | Grouping of positive and negative emotions | These clusters when coupled with emotion mimicry may result in better empathetic responses | [58] |
| Emotion Causes | Specific words in the user message that carry the causes of a particular emotion | Emotion causes go beyond emotions to understand the user's issues and feelings in a more nuanced way | [26], [60], [61] |
| Dialog Act/Intent | Utterance in the context of a conversational dialog that serves a function | Understanding the user's intent is importing for charting the next steps in the dialog process | [27], [28], [30] |
| Persona | The social face an individual presents to the world | Persona is highly correlated with personality which in turn influences empathy | [25], [62]–[64] |
| Verbal and Non-verbal Behavior | Expressions of emotion in both textual and non-textual (video, audio) data | Multimodal detection of emotion goes beyond the current dominant unimodal approaches in fully understanding the user's emotion state | [65] |
| Communication Mechanism | The higher-level and abstract factor relating to empathy expression | Can produce better empathetic responses when integrated with other factors (dialog acts) | [30] |
| External Knowledge | Commonsense and emotion lexicon knowledge bases queried with the words from user messages | Helps with making meaningful inferences about the user's emotional state | [66]–[69] |
| ECS | | | |
| Multiple Listeners | Listeners dedicated to different emotion classes | Each listener is optimized to react to the particular emotion so that more specific empathetic responses could be generated | [59] |

| Response Types | Preset response templates differ by the context of the user messages | May help in situations where the conversations are not required to be lengthy and engaging | [70] |
|---|---|---|---|
| Anticipated Emotion | Predicting the future emotional state of the users based on the generated ECS response | Helps in minimizing the divergence between the anticipated emotions and ground truth emotion distributions. Reciprocal altruism can also be achieved | [71], [72] |
| Anticipated Sentiment | Predicting the future sentiment of the users based on the generated ECS response | May help in ensuring satisfactory sentiment polarity in future user messages | [73], [74] |
| ECS Messages and User | | | |
| Exemplars | Small text snippets that have been added to the ECS responses | Enforces the presence of empathetic reactions in the ECS responses | [29] |
| Affective Matching Mechanisms | Matching or mirroring the emotional state of the user | When emotional consensus is reached between the user and ECS agent, the quality of empathetic conversations seems to improve | [58], [75] |
| User Feedback | Feedback about the responses generated by the ECS agents | Through either active or adversarial learning, the conversations are tailored according to user feedback so that more effective empathy is exhibited | [62], [63], [76], [77] |

*(d) Emotion Causes*: Emotion causes refer to the specific words in the user message that carry the causes of a particular emotion. Studies that have identified the emotion causes have generated more relevant and empathetic responses from the ECS. This area has been an active area of research due to promising improvements [26], [60], [61]. For example, for the sentence "I really wanted to get the part in the school rendition of a play. But my friend got it", the emotion causes words are "rendition", "friend" and "got". The approach of extracting causes for particular emotions in natural language text applies to open-domain settings i.e., chit-chat context. For domains such as customer service, another related research topic known as aspect-based sentiment/emotion analysis [78], [79] is more suitable as specific aspect terms and aspect descriptions or opinion terms need to be extracted from text.

*(e) Dialog Act/ Intent*: Dialog act and intent classification is a mandatory task in task-based and goal-oriented dialog systems [80]. For instance, when the user is seeking the help of a CAI agent for booking a flight ticket. The agent will have to detect the user intent as "book_ticket" so that the corresponding information could be collected from the user for completing the ticket booking. This task is not usually performed in open-domain dialog systems. Although ECS studies predominantly follow open-domain dialog systems architecture, few ECS studies have attempted to classify the dialogue act/ intent of conversations. In two ECS studies [27], [28], new intents have been identified from the EMPATHETICDIALOGUES dataset [24]. Eight intents were identified specifically. They are questioning, agreeing, acknowledging, sympathizing, encouraging, consoling, suggesting, and wishing. The eight new intents are combined with the existing 32 emotion classes along with a neutral class to form a hybrid emotion-intent category set. The authors opine that datasets used in ECS studies need to be annotated with these new emotion-intents to facilitate better response generation. This emotion-intent category data is utilized in another ECS study [30] to emphasize its usefulness.

*(f) Persona*: Persona-based research in CAI agents mandates the agent to showcase a particular personality while interacting with users [81]. Persona is highly correlated with personality which in turn influences empathy as per research [82], [83]. Certain ECS studies have followed this persona-based approach for improving the empathetic response generation performance [25], [62]–[64]. A Persona-based Empathetic Conversation (PEC) dataset [25] has also been published as a part of this research. The findings from these ECS studies indicate that persona has a greater impact on empathetic conversations than non-empathetic conversations.

*(g) Verbal and Non-verbal Behavior*: Multimodal emotion recognition or affect detection has gained considerable attention in affective computing, where the combination of features from language, facial expressions, and tone of voice are found to perform generally better than single-modality [84]–[86]. However, multimodal approaches are a rarity in ECS studies. As a part of this review, we were able to identify one study [65] where multimodal inputs were processed. The dataset for this study was prepared by annotating an existing human-agent video interview. Both verbal (text-based) and non-verbal (video-based) behaviors were identified to understand the user's emotions for response generation.

*(h) Communication Mechanism*: Communication mechanism refers to a higher-level and abstract factor relating to empathy expression. It was introduced in the study where the objective was to detect empathy in the text format [87]. There are three types of communication mechanisms namely emotional reactions, interpretations, and explorations. The emotional reactions are recorded at two levels – weak communication ( little expression of the felt emotion) and strong communication (full expression of the felt emotion). Similarly, interpretation is also recorded at weak (lower cognitive understanding of the speaker's emotional state) and strong (higher cognitive understanding of the speaker's emotional state) levels. The explorations mechanism is for improving the understanding of emotions by exploring the feelings not directly stated by the speaker. A weak exploration is one where a generic question is asked to the speaker (e.g., "what happened?") while a strong exploration indicates stronger empathy (e.g., "Are you feeling alone right now?"). This communication mechanism concept has been used in an ECS study as one of the input factors [30]. The study highlights the usefulness of this factor in generating empathetic responses.

*(i) External Knowledge*: Entity extraction from user messages is not a common activity in existing ECS studies. There are a few studies that have utilized this activity [66]–[68]. In these three studies, the entities and the relations between the entities have been extracted from preexisting knowledge stored in external knowledge bases (PolarisX in [66], ConceptNet in [68], [69], and ATOMIC in [67]) to make meaningful inferences. ConceptNet [88] is queried in two studies [68], [69] to extract both additional concepts and relations. Let us consider two examples. In the first example from [68], "I was hiking in the outback from Australia the other day", the word "hiking" is related to the words "enjoy nature" through the relation "has_subevent" in ConceptNet. In the second example from [69], "I started to cough blood 3 days ago", the word "cough" is related to the word "illness" in ConceptNet. Compared to ConceptNet, PolarisX can consistently collect new data from social media and the web [89]. In the ECS study [66], it is used to establish relations between the words (concepts) in the user messages. ATOMIC is a knowledge base of commonsense reasoning inferences about everyday if-then events [90]. It is used in [67] to infer six commonsense relations for the person involved in an emotional event. The usage of such knowledge bases in ECS studies helps in understanding the speaker/user's situation in a more detailed way by inferring implicit information which is latent in the messages.

## ECS

In existing studies, four concepts have been analyzed at the ECS component level (refer to Fig. 2). It is to be re-iterated that the responses from the ECS models are usually generated (using natural language generation models) or retrieved (using templates or responses database).

*(a) Multiple Listeners*: In one study [59], response prediction is performed by combining the responses from multiple listeners (response generation models) dedicated to different emotion classes. The rationale is that each listener is optimized to react to a particular emotion. At the final level, a meta-listener softly combines the response from all the listeners. This approach did not yield better offline evaluation results but scored higher in user evaluation.

*(b) Response Types*: In another study [70], the concept of preset response types was explored. In this study, three types of responses will be provided back to the users based on their corresponding utterances. The response types are empathetic, context-based, and intent-based. This approach was prototyped with a depression questionnaire chatbot which interacts with the users to get responses. The chatbot's empathetic responses included coping mechanisms depending on the answers to the particular questions. We opine that this approach is simplistic as preset responses are not scalable. Also, the response types need to be hybrid in most cases as empathy needs to be embedded in the chatbot responses whenever required.

*(c) Anticipated Emotion:* In ECS studies, the empathetic response generation process generally takes the emotion of the current user message or the historical user messages into consideration. The anticipation of the next emotional reaction of the user is taken into the response generation study in two ECS studies. In [71], this approach helps in minimizing the divergence between the anticipated emotion and ground truth emotion distributions. However, this approach has not helped in producing better results for all the considered evaluation metrics. Hence, there is scope for improvement. In another study [72], multiple chatbots are trained to achieve reciprocal altruism based on the rationale that altruistic behaviors are exhibited by humans during interactions. In this study, a reinforcement learning model is used to predict the next emotion of the user. Along with the future emotion, a conceptual human model has also been additionally implemented to simulate future responses from the users.

*(d) Anticipated Sentiment*: Similar to anticipated emotion, there have been a few studies that have attempted to predict or anticipate the future sentiment of the users [73], [74]. In [73], the anticipated sentiment concept is referred to as sentiment look-ahead, which is a reward function under a reinforcement learning framework. This framework provides higher rewards to the response generation models when the generated response improves the user's sentiment. In another related study [74], the anticipated sentiment is referred to as the affect label. Two new modules namely the discrimination and re-writing modules are used in this study in place of the reinforcement learning framework in the other ECS study [73]. The discrimination module decides whether the generated response will fetch satisfactory polarity. If the condition is not met, the rewriting module rewrites the response so that the sentiment polarity is improved.

## ECS Messages

At the ECS messages level, there have been a few concepts that have been analyzed.

*(a) Exemplars*: Exemplars are small text snippets that have been added to the ECS responses so that the overall messages appear more empathetic [29]. These exemplars are also retrieved from the same training set which is used for the main model training. Examples of exemplars are "why what happened", and "oh no, why is that". These snippets are considered exemplary relevant responses for particular contexts. In the overall architecture, an exemplar injection module sends an exemplar to the final response general model which includes the exemplars as a prefix to the final generated response. Although this approach shows promise, it was found to need improvement in multiturn contexts where the user and agent have back-and-forth conversations.

*(b) Affective Matching Mechanisms*: The affective matching mechanism has been experimented with in a few ECS studies in which the user messages and ECS messages are analyzed in a combined manner. This mechanism has been operationalized through emotional mimicry [58] and emotional consensus [75]. In [58], the empathetic responses are made to mimic the emotion of the user while accounting for their sentiment. This study was discussed earlier under the positive and negative emotion cluster sub-section in the User Messages component. In [75], the affective matching mechanism is grounded on emotional consensus since empathy is triggered when the user and the agent link similar experiences with their emotions converging. For this purpose, the study makes use of a dual-generative model to construct emotional consensus. Based on the evaluation results, this approach seems to provide better results on all evaluation metrics when benchmarked against the SOTA approaches. Hence, this emotional consensus based ECS model shows promise for future ECS studies.

(c) *User Feedback*: The user's feedback has also been sought in a few studies to improve the response generation capabilities of ECS dynamically. This feature has been achieved through active learning [62], [63] and adversarial learning [76], [77]. Through the active learning approach used in [62], [63], the user can provide two types of optional feedback to the ECS agent. The first feedback is about reporting unethical messages while the second feedback is the expected or correct empathetic response that the ECS agent could have provided. Other than the learning feature, the ECS model used in these studies also incorporates a persona-based finetuning so that the agent comes across with a suitable persona. In the demo website of this chatbot [91], emojis are embedded for both user and chatbot responses, although the rationale for emoji inclusion has not been provided in the paper. The model shows promising results in offline evaluation, but human evaluation experiment results have not been reported. The adversarial learning framework is implemented in two ECS studies [76], [77]. This framework is used for analyzing user feedback to identify whether the generated responses evoke adequate emotion perceptivity. Two discriminators (semantic and emotional) are used for this purpose to handle user feedback. In both offline and human evaluation experiments, this approach produced better results for most metrics. Hence, this adversarial learning approach can also be considered as a promising method for future ECS studies.

## 4.2 Datasets

In Table 3, the datasets used in ECS studies have been listed. Most of the datasets were created specifically for ECS studies while some datasets have been adopted from earlier general-purpose CAI dialog systems studies.

### TABLE 2
### DATASETS USED IN ECS STUDIES

| SL.no | Dataset | Method | Studies # | Studies |
|---|---|---|---|---|
| Created for ECS Studies | | | | |
| 1 | EMPATHETICDIALOGUES [24] | Crowdsourcing | 26 | [24], [29], [30], [53], [58]–[63], [67]–[69], [71]–[73], [75]–[77], [92]–[98] |
| 2 | Persona-based Empathetic Conversation (PEC) [25] | Base data extracted from Reddit | 2 | [25], [30] |
| 3 | ArabicEmpathicDialogues [94] | Crowdsourcing | 2 | [93], [94] |
| 4 | XiaoAI Empathetic Conversation (X-EMAC) dataset [26] | Base data extracted from XiaoAI online logs | 1 | [26] |
| 5 | Empathetic Open-Subtitles dialogues dataset [27] | Base data used from OpenSubtitles dataset | 1 | [27] |
| 6 | Emotional dialogues in OpenSubtitles (EDOS) dataset [28] | Base data used from OpenSubtitles dataset | 1 | [28] |
| Adopted from other CAI studies | | | | |
| 7 | DailyDialog [99] | Base Data Extracted from Multiple Websites | 5 | [66], [73], [97], [98], [100] |
| 8 | BookCorpus [101] | Base Data Extracted from Books | 3 | [62], [63], [100] |
| 9 | PersonaChat [102] | Crowdsourcing | 3 | [62], [63], [73] |

### Datasets created for ECS Studies

The seminal EMPATHETICDIALOGUES dataset [24] has been used for evaluation directly or indirectly in 26 studies thus far. It contains 24,850 conversations. This dataset was prepared through a crowdsourcing approach where 810 participants played the role of a speaker or listener. They were instructed to respond for at least six turns empathetically. First, the speaker was provided with an emotion class (a total of 32 emotion classes were used), and he/she was advised to speak about a real-life scenario corresponding to the assigned emotion class. Next, the speaker was paired with a listener, and they proceed to converse about the real-world scenario. Both the speaker and listener are not made aware of the assigned emotion class. Each conversation is restricted to 4-8 utterances. A sample dialogue for the emotion class "afraid" is provided from the dataset in Figure 3. In subsequent studies, the EMPATHETICDIALOGUES dataset has been further augmented with eight intent categories [103] and has been separately translated to other languages (e.g., Arabic [93], [94]).

**Emotion Class: Afraid**

**Situation:** Last year a tree fell on my house while my family was at home. The tree broke through the ceiling just a few feet away from my daughter. The experience was terrifying.

--------------------------------------------------

**Speaker:** What a difference a year makes. Last year one evening my family was at home when a tree fell on the house and broke through the ceiling.

**Listener:** That's very scary. I hope no one got hurt.

**Speaker:** We were OK, though the tree broke through only a few feet away from my daughter.

**Listener:** So happy everyone was fine!! Everything else can be fixed.

**Speaker:** Indeed. We were out of the house for five months while repairs were being done, but now the house is better than ever.

**Listener:** So good to hear. Might want to trim some trees lol

Fig 3: Sample Conversation from EMPATHETICDIALOGUES Dataset

The second most popular dataset is the Persona-based Empathetic Conversation (PEC) dataset [25] comprising 355,000 conversations. It was prepared with the source data extracted from Reddit, specifically from two subreddits *happy* and *offmychest*. The *happy* subreddit is a forum where users share happy stories and thoughts from their personal lives whereas the *offmychest* subreddit forum is used by users for sharing emotional anecdotes. To prepare the dataset, crowdsourced workers annotated the source data extracted from the two aforementioned subreddits with appropriate empathy labels. Three annotators were involved in the process. The inter-annotator agreement values for sentiment and empathy labels were 0.725 and 0.617, respectively. The uniqueness of the PEC dataset is that it can be used to train ECS models for exhibiting both persona and empathy. This dataset has been used by a later published ECS study [30].

The XiaoAI Empathetic Conversation (X-EMAC) dataset [26] comprises 16,873 conversations. It was prepared by extracting the base data from XiaoAI online logs. The base data was annotated with four emotion classes (sad, anger, joy, and others). Next, psychologists worked on this annotated dataset to create response templates. These responses were formulated based on counseling strategies of active listening and effective questioning. The templates were designed specifically for each emotion class. These templates were next provided as responses to those user queries that are classified into certain emotion classes (sad, anger, joy), and the next turn of real-time responses from the users was subsequently collected. For these three turns (user-template-user), annotators identified the emotion causes span and also wrote empathetic responses. There is no information provided about the inter-annotator agreement in this paper.

Two related empathetic conversational datasets, namely the Empathetic OpenSubtitles Dialogues dataset [27] and Emotional Dialogues in OpenSubtitles (EDOS) dataset [28], were prepared with base data extracted from the public OpenSubtitles dataset [104]. The EDOS dataset comprises one million emotional conversations from movie subtitles. Each turn is annotated with 41 emotion classes. This dataset was created to launch a comprehensive empathetic dataset that is bigger than the EMPATHETICDIALOGUES dataset [24]. The authors argue that movie dialogues closely mirror real-world conversations and hence can be considered to be a suitable substitute. In this study, automatic data augmentation techniques were used to expand a manually annotated set of 9K movie dialogues to the full one million emotional dialogues.

Another dataset called EMPATHETICPERSONAS [64] was created through crowdsourcing. This small-scale dataset of 3,324 conversations was constructed through survey responses where the respondents were asked to perform two tasks – *1) respond with emotions to the question – How are you feeling? and 2) rewrite a set of base utterances to render them empathetic.* No further information has been provided about the additional instructions that were provided to the respondents. Based on the provided information, the survey respondents seem to have been provided with enough leeway to rewrite utterances as per their own judgement. From the overall dataset, there are 2,143 empathetic rewritings of 45 base utterances. Two annotators were recruited to annotate the rewritings with different levels of empathy using a 3-point empathy scale (0: non-empathy, 1: weak empathy, and 2: strong empathy). No information is provided regarding the inter-annotator agreement.

### Datasets adopted from general CAI dialog studies
Apart from the empathetic conversational datasets, generic conversational datasets have also been used for pretraining, training, and evaluation of response generation modules in ECS studies. The DailyDialog dataset [99] is the popular dataset in this category, followed by BookCorpus [68], PersonaChat [102], and Douban Conversation Corpus [105]. These generic datasets are mainly meant to enrich the conversation quality and make the conversations realistic.

The DailyDialog dataset [99] comprises 13,118 multi-turn dialogues about daily life conversations with the source data extracted from various websites. The dialogues are annotated with two purposes – "exchanging information" and "enhancing social bonding" and four dialog acts (inform, questions, directives, commissive). This dataset is used in ECS studies for pretraining the response generation model in [73], [100]; training the emotion classifier and response generation model in [97] and [98] respectively; and for evaluation in [66].

The next popular dataset is the BookCorpus [101] dataset was constructed from 11,038 books, containing a total of approximately 74 million sentences [106]. The books are from different genres such as romance, fantasy, and science fiction. Since the dataset has extensive coverage of words, it has been used for only pre-training purposes in ECS studies [62], [63], [100].

The PersonaChat [102] dataset comprises 162,064 conversations between crowdsourced participants who were randomly paired. The participants were asked to act the part of a given persona. They were requested to chat naturally and get familiar with each other during the conversation. 1,155 personas were considered for this dataset. Each persona is represented by at least five sentences that given an indication about the particular persona. Similar to the BookCorpus dataset, this dataset has also been used solely for pretraining in ECS studies [62], [63], [73].

### 4.3 Techniques used for Request Processing and Response Generation
This section focuses on the techniques or algorithms employed in ECS studies for both request processing and response generation. In CAI system architectures, there are multiple modules such as NLP pre-processing, natural language understanding (NLU), dialogue management, and response generation [107]. Nevertheless, these modules are not utilized together in ECS studies since the focus of these studies is to develop models to showcase empathy in an open-domain setting. Since most of these studies are centered around the EMPATHETICDIALOGUES dataset [24], the emotion detection module operates before the response generation phase to identify the emotion class in the user message. In certain studies [53]–[55], a sentiment classifier is placed instead of an emotion classifier to ascertain the sentiment class. While different deep learning architectures are used to classify emotion/sentiment, the classifiers are tuned with pre-trained models. BERT [108] and RoBERTa [109] are the popular models for this pre-training purpose, with usage in four studies. Deepmoji [110] and VADER [111] algorithms have been used in one study for emotion and sentiment identification, respectively. Apart from emotion/sentiment detection, other features are extracted or identified in user messages (requests). These features have been presented in Section 4.1. In this review, the algorithms or techniques used for such feature extraction are not presented as they have not emerged as standard procedures in ECS implementations.

There are three types of response generation methods in CAI systems; rule-based, retrieval, and generative [4]. The rule-based method is the fastest to initialize and deploy but requires constant monitoring and editing. The retrieval-based response generation method retrieves the most relevant response from a database of predefined responses [112]. However, the retrieval-based method is not the dominant response generation method in ECS studies, and it has been used in a handful of studies [24], [74], [113]. On the other hand, the generative-based method is the popular method in ECS studies, mirroring the latest trend in dialogue systems. Generative models can produce new dialogues based on large amounts of conversational training data [4]. The standard transformer model [114] is the most frequently used model in response generation modules in ECS studies, followed by GPT [115], Seq2Seq [116], and GPT-2 [117]. The Text-To-Text Transfer Transformer (T5) [118], one of the advanced latest models providing the best results in multiple NLP tasks, is gaining prominence in ECS studies with usage in three studies. The BERT [108] model and its different variants, such as CoBERT [25] and ALBERT [119] have also been used for creating response generation models. DialoGPT [120] model, which is considered state-of-the-art in pre-trained response generation models, does not seem to poplar among ECS studies with deployment in a single study [92]. It is perceivable that ECS studies have employed the latest deep learning architectures in both request processing and response generation modules. The techniques, or models used for emotion/sentiment detection and response generation in ECS studies, are listed in Table 3.

TABLE 3
REQUEST PROCESSING AND RESPONSE GENERATION TECHNIQUES IN ECS STUDIES

| Module | Model/Technique | No. of Studies | Studies |
|---|---|---|---|
| Emotion/Sentiment Classification | BERT [108] | 4 | [24], [71]–[73] |
| | RoBERTa [109] | 4 | [27], [30], [64], [68] |
| | Deepmoji [110], VADER [111] | 2 | [29], [97] |
| Response Generation | Standard Transformer [114] | 12 | [24], [27], [28], [53], [58]–[60], [67], [75]–[77], [98] |
| | GPT [115] | 7 | [26], [62], [63], [66], [72], [97], [100] |
| | Seq2Seq [116] | 5 | [73], [93], [94], [96], [113] |
| | GPT-2 [117] | 3 | [30], [68], [71] |
| | Text-To-Text Transfer Transformer (T5) [118] | 3 | [29], [53], [64] |
| | Others (DialoGPT [120], CoBERT [25], BERT [108], ALBERT [119]) | 5 | [25], [66], [72], [74], [92] |

## 4.4 Evaluation Approaches and Metrics

Evaluation results were reported in 37 ECS studies surveyed. Offline evaluation is the most-used evaluation approach in ECS studies, with 34 of these studies using the approach. In offline evaluation, the ECS techniques are typically evaluated based on their ability to reproduce the listener's responses to a speaker (as designated in the training dataset). In ECS architectures with an emotion/sentiment detection module, Accuracy is the primary metric used for evaluating the classification performance.

For the main task of evaluating the response generation ability, Perplexity is the most popular metric used in 22 studies, followed by BLEU [121] in 18 studies. Perplexity is a model-dependent metric that measures how well a probability model predicts a given sample. On the other hand, the Bilingual Evaluation Understudy Score (BLEU) compares the generated response against the gold standard (the actual response). Beyond ECS studies, BLEU is the most frequently used evaluation metric in conversational dialogue systems [31], [122]. The third most popular metric is the Distinct-n (dist-n) metric with 11 studies. This metric enumerates the percentage of unique n-grams in the responses [123]. Typically, unigrams (dist-1), bigrams (dist-2) and trigrams (dist-3) are reported. The other prominent offline evaluation metrics are Sentence Embedding Similarity, F1, Mean Reciprocal Rank (MRR), Diversity, Loss, and Recall. Ablation study does not seem to be popular with ECS offline evaluation experiments, with only nine papers reporting ablation evaluation results.

In 26 ECS studies, user evaluation results have been reported. In these studies, human ratings are collected for specific user perception-related metrics. The three most popular user metrics are Empathy, Relevance, and Fluency measures. For the Empathy measure, the related question is *"Did the responses show understanding of the feelings of the person talking about their experience?"*. The question for the Relevance measure is *"Did the responses seem appropriate to the conversation? Were they on-topic?"* and the question for the Fluency measure is *"could you understand the responses? Did the language seem accurate?"*. These three metrics were first introduced in the EMPATHETICDIALOGUES study [24]. There have a few other user metrics used in ECS studies. Net Sale Value (NSV) was proposed in [26] to measure the preference towards a particular ECS implementation by using the number of upvotes (likes) and downvotes (dislikes). The formula is (#upvotes-#downvotes)*(#upvotes+#downvotes). In [67], Coherence and Informativeness are used along with the Empathy metric. The questions for these two metrics are: *which response is more coherent in content and relevant to the context* and *which response conveys more information about the context*? Human A/B testing has been performed in five studies [29], [58], [60], [75], [113]. In A/B testing, human annotators are requested to pick the model with the best response for each of the sub-sampled test instances for two models, A and B.

## 4.5 Notable State-of-the-Art (SOTA) Approaches in ECS Studies

In this section, we discuss most notable SOTA approaches in ECS studies.

First, the study of the EMPATHETICDIALOGUES dataset [24] gains the highest popularity in ECS studies, and it mandates a discussion on the response generation models proposed in the corresponding study. Three additional studies have been handpicked by the authors based on three criteria namely novelty, comprehensiveness, and performance. Novelty pertains to the impact and level of usage of empathy-related concepts which are entirely novel or operationalized from conceptual empathy models and frameworks. Comprehensiveness refers to extent of the ECS evaluation, i.e., whether the proposed model has been evaluated through offline evaluations, human evaluations, and ablation studies. The criteria also consider whether a particular study has benchmarked the proposed model(s) against the existing baseline approaches. Finally, the performance criteria provide weightage to studies that have reported better evaluation results for the considered metrics.

### Empathetic Open-domain Conversation Models

As introduced in section 4.2, the EMPATHETICDIALOGUES dataset comprises conversations grounded in specific situations in which the speaker and listener converse with each other. The response generation models are designed to emulate the role of the listener who responds empathetically. This role has no prior knowledge of the emotion class and the grounded situation.

With a dialogue context $X$ of $n$ past conversation utterances that are concatenated and tokenized as $\{x_1, x_2, x_3 \ldots x_m\}$, the models are trained to produce a response $Y$ by maximizing the likelihood $p(Y | X)$. In this study, both retrieval-based models and generative models are proposed. For this retrieval-based setup, the model is provided with a large number of potential responses and the task is to select the best suitable response i.e., When two input vectors $h_x$ and $h_y$ are provided, the best response candidates maximize the dot product of $h_x$ and $h_y$ where $h_x$ is the encoded data of the historical utterances $\{x_1, x_2 \ldots, x_m\}$, and $h_y$ is the encoded data of the possible responses $\{y_1, y_2, \ldots, y_m\}$. The specific transformer-based architecture from [124] and BERT-based [108] architectures are experimented with for retrieval-based setup. Thereby, there are two retrieval-based architectures proposed in this study.

In the case of a generative model, the standard transformer [114] architecture is used. The encoder is the same as the earlier models while the decoder takes the encoder outputs for predicting a sequence of words that combine to form the response $y$. The model is trained to minimize the negative log-likelihood of the response $\underline{y}$.

For the three proposed architectures in the study, multiple types of training options along with different data sources are evaluated. In addition, the authors also experiment with the inclusion of information from external predictors to check if the overall performance improves. In Table 4, the 21 different model variations are listed based on these three dimensions. For the training options, the variations experimented with are pre-trained and fine-tuned. For the response data sources, three datasets are considered. They are Reddit (R), DailyDailog [99] (DD) and EMPATHETICDIALOGUES [24] (ED). For the external predictors, Emoprepend-1 and Topicprepend-1 are considered. Emoprepend-1 is an emotion classifier that predicts the emotion class from the situation description text. This classifier is trained on the EMPATHETICDIALOGUES dataset itself. Topicprepend-1 is a classifier to predict topics represented by the text. This classifier is trained on a newsgroup dataset [125].

TABLE 4
MODEL VARIATIONS IN EMPATHETICDIALOGUES ECS STUDY

| Model | Response Source | Architecture | | |
|---|---|---|---|---|
| | | Retrieval | Retrieval with BERT | Generative |
| Pretrained | R | M1 | M2 | M3 |
| | ED | M4 | M5 | M6 |
| Fine-tuned | ED | M7 | M8 | M9 |
| | ED+DD | M10 | M11 | M12 |
| | ED+DD+R | M13 | M14 | M15 |
| Emoprepend-1 | ED | M16 | M17 | M18 |
| Topicprepend-1 | ED | M19 | M20 | M21 |

Out of the 24,850 total conversations in the dataset, 19,533 conversations are allotted to the training set while 2,770 and 2,547 conversations are allotted to the validation and test sets respectively. Both offline and user evaluation experiments are conducted in this study. Three offline evaluation metrics are precision, average BLEU, and perplexity. Compared to pre-trained and external predictor model variations, the fine-tuned models (M7, M8, and M9) with EMPATHETICDIALOGUES dataset as the response source provide the best results for retrieval and generative architectures. The same models also provide the best results in the user evaluation where the metrics empathy, relevance, and fluency are first introduced in this study. For the retrieval-based architecture with BERT, topicprepend-1 models (M20) provide the best results, thereby indicating this approach might not boost the performance of smaller models but rather benefit the bigger models. The study results prove the utility of the EMPATHETICDIALOGUES dataset over the earlier published datasets and also underline the effectiveness of fine-tuning in both retrieval-based and generative models.

*Emotion Causes Oriented Empathetic Response Generation Models*

Emotion cause is one of the promising empathy-related concepts in ECS studies since empathetic response generation is centered on particular words in the user text that allude to the cause behind the user's emotion. Out of the three ECS studies where emotion cause-oriented response generation models were proposed, we have considered one study [60] which satisfies our three criteria.

In this study's main model, there are two components emotion reasoner and response generator. The emotion reasoner predicts the emotion class when provided with a dialogue context and also identifies words that convey the cause for that particular emotion class. The dialogue context comprises of $\{u_1, u_2, \ldots u_n\}$ of $n$ utterances, and each utterance $u_i$ comprises of words $\{w_1^i, w_2^i \ldots w_m^i\}$ of $m$ tokens. The input sequence $X$ $\{x_1, x_2, \ldots x_n\}$ is formed by concatenating all the utterances that are separated by a special [SEP] token. For this input sequence $X$, the response generator is expected to produce an appropriate empathetic response $Y$ $\{y_1, y_2, \ldots y_n\}$.

In the emotion reasoner, emotion prediction is the first task that is executed by a transformer encoder. This task is considered a text classification task. A representation of the words in the dialog context is fed into the encoder. Based on this data, context emotion distribution is computed. For the second task of emotion cause detection, an existing model [126] is leveraged. This task is considered a sequence labeling task. This is accomplished by labeling each word in the input sequence with an emotion cause label $\{0,1\}$.

The predicted emotion class and emotion cause words are input into the response generator. The standard transformer [114] is used as part of this module. Trainable emotion embeddings are used so that the 32 emotion class labels could be represented. Each word in the input sequence is represented as a summation of word embedding, positional embedding and emotion embedding. First, the input sequence $X$ is fed into the encoder section of the transformer so that contextualized word representations are made available for the decoder. The emotion cause words are forced into the response generation process through a gated attention mechanism whereby a sequence of gates $G$ $\{g_1, g_2, \ldots g_n\}$ dynamically select elements that are related to the emotion cause words. With these gates, two strategies namely hard gating and soft gating are explored to vary the impact of the emotion-cause words on the response generation process. The former is a strict strategy while the latter is a comparatively flexible strategy.

For the evaluation experiment, the EMPATHETICDIALOGUES dataset [24] is used. The dataset is split into training, testing, and validation subsets with an 8:1:1 split. The proposed model has two variations namely soft and hard, referring to the soft gating and hard gating approaches respectively. The baselines selected for this study are MoEL (Mixture of Empathetic Listeners) [59], MIME (MIMicking Emotions for Empathetic Response Generation) [58], EmpDG (Empathetic Dialogue Generation) [77] and MK-EDG (Multiple Knowledge Empathetic Dialogue Generation) [69]. These baseline models have been covered in this review. The offline evaluation metrics used in this study are BLEU, distinct-1/distinct-2, BERTscore [127], and accuracy (for emotion prediction). The constructs, namely empathy, relevance, and fluency, are employed for the user evaluation. Additionally, an ablation study and human a/b testing were conducted. For all the offline evaluation metrics, both the hard and soft models provided the best results. The ablation-based analysis indicated the importance of emotion prediction and emotion cause extraction in the overall proposed model's performance with the latter boosting the model's performance. The user evaluation study and human a/b testing results also underline the better performance of the proposed models when compared to the baselines.

*External Knowledge for Empathetic Response Generation Models*

As highlighted in section 4.1, ECS architectures can benefit from external knowledge sources such as knowledge bases since commonsense and emotion-related knowledge can help in inferring latent concepts and relations from user messages. As a part of this review, we identified four papers in which knowledge bases were used. We selected the Knowledge-aware EMPathetic response generation method (KEMP) study [69] since it satisfies the three criteria. The premise for this study is that there is not much overlap found between dialogue history and response messages at the non-stopword word level in the popular EMPATHETICDIALOGUES dataset [24] which is used in multiple ECS studies. However, when external knowledge is incorporated, the authors claim that the ECS agents can obtain useful hints from the dialogue history so that better empathetic responses could be generated.

The proposed KEMP method in this study incorporates three components which are an emotional context graph, emotional context encoder, and emotion-dependency decoder. Two knowledge bases ConcepetNet [88] and NRC_VAD [128] are leveraged. ConceptNet [88] is used to extract a set of candidate tuples of the structure {*head concept; relation; tail concept; confidence score*} for the non-stopword words in the user messages. An example given in the paper is {*birthday; RelatedTo; happy; 0:19*}. NRC VAD [128] is a lexicon of VAD (Valence-Arousal-Dominance) vectors for around 20K English words. In this study, NRC_VAD is used to score the emotion intensity for the dialogue words and the external concepts extracted from ConceptNet. The concepts with higher emotion intensity are fed into the KEMP model.

The aforementioned three components of KEMP are executed sequentially thereby forming three phases. In the first phase, the dialogue history $D$ $\{x_1, x_2, \ldots x_n\}$ is enriched with the external knowledge to form the emotional context graph $G$. In the second phase, emotional signals $e_p$ of $D$ are distilled based on the embedding and the emotion intensity data (obtained from NRC_VAD). In the third phase, the emotional signals $e_p$ and graph $G$ are used to learn emotional dependencies using a cross-attention mechanism. In the last step, the empathetic responses $Y$ $\{y_1, y_2, \ldots y_n\}$ are generated. A multi-task learning framework is used in KEMP to minimize the losses.

Similar to the previous SOTA approach, the evaluation experiments are conducted on the EMPATHETICDIALOGUES dataset [24]. The dataset is split into the training set with 17,802 dialogues, the validation set with 2,628 dialogues, and the testing set with 2,494 dialogs. The KEMP method is compared with five baseline approaches. They are standard transformer [114], Emoprepend-1 from [24], MoEL [59], MIME [58], and EmpDG [77]. The offline evaluation metrics used in this study are accuracy (for emotion prediction), perplexity, BLEU, and distinct-1/distinct-2. The constructs empathy, relevance and fluency are utilized for the user evaluation. Ablation study and human a/b testing results were also reported in this paper. The offline and user evaluation results indicate the superiority of the KEMP method in outperforming all the baseline approaches. The only exception is the fluency metric where EmpDG method produces slightly better results. In the ablation study, the emotional context encoder and emotion-dependency decoders are replaced with standard transformers to ascertain the effectiveness of these two components. The study results indicate that the performance of the model deteriorates without these components, thereby demonstrating the utility of external knowledge and modeling emotional dependencies. Although this KEMP model provides better performance than the earlier baseline approaches, the emotion causes model [60] (discussed in the previous sub-section) provides even better performance, thereby highlighting the usefulness of emotion cause extraction in ECS studies.

*Empathetic Response Generation based on Affect Matching*
Affect matching through mimicry is one of the basic empathic behaviors in conceptual empathy models, which has been highlighted in Section 3. There are two ECS studies [58], [75] that have employed the affect matching mechanism in the model architecture. We selected the study with the better affect-matching design and evaluation results. In the selected study [75], the authors propose an approach to establish "emotion consensus" between the user (speaker) and the ECS agent (listener). They argue that empathy is triggered when two interlocutors attempt to link similar experiences and their emotions converge on the same point, as a result. The model proposed in this study generates pseudo-empathetic conversations for unpaired emotional data in the EMPATHETICDIALOGUES dataset [24] since a bidirectional process is necessitated for establishing emotion consensus.

To realize the study objectives, a dual-generative model (DualEmp) is proposed. It concurrently constructs emotion consensus and utilizes unpaired data from EMPATHETICDIALOGUES dataset. DualEmp unites a forward dialogue model (generating a response based on its context) and a backward dialogue model (generating a context based on its responses) with a discrete latent variable. With context $C$ = $\{S_1, L_1, S_2, L_2 \ldots \ldots S_i\}$, where $S_i$ denotes speaker and $L_i$ denotes listener, the goal is to track the speaker's emotional state from $C$, and generate an empathetic response $Y$. As per the architecture, the DualEmp has five modules comprising the forward encoder $f_{enc}$, forward decoder $f_{dec}$, backward encoder $b_{enc}$, backward decoder $b_{dec}$, and discrete latent variable $z_e$. The variable $z_e$ is inferred from both $C$ and $Y$ and it is used to capture emotion consensus shared in each $\{C, Y\}$ pair in the dataset.

The forward and backward dialogue models have the same architecture. In the encoder, all the utterances in $\underline{C}$ are concatenated into a long sequence. For each token $w$ in $C$, three embedding spaces are summed up. They are word embedding space, positional embedding space, and role embedding space where the role is either speaker or listener. A standard transformer encoder [114] is used on top of the summed embedding space to get the context representation. After the encoder, the discrete latent variable $z_e$ is used to capture the emotion consensus between $C$ and $Y$. This is performed by training a classifier using the cross-entropy loss between the embedding space of $z_e$ and ground truth emotion label $e$. Finally, the decoder model focuses on emotion consensus with the application of an emotion-enhanced attention mechanism in the cross-attention layer of the transformer decoder i.e., the embedding space of $z_e$ is first concatenated with the decoder input to get representations of $Y$. Next $Y$ is fed into forward decoder $f_{dec}$. The DualEmp model can be trained with both paired data and unpaired data from the dataset

Similar to the previous two SOTA ECS models, DualEmp model has been benchmarked against multiple baseline approaches. The five baselines considered are the Emoprepend-1 model from EMPATHETICDIALOGUES study [24], MoEL [59], MIME [58], EmpDG [77], and DualVAE [129] model which is a dual decoder model rarely used for benchmarking in ECS studies. Three variations of the DualEmp model are used for evaluation. They are (1) a single-paired variation with only a forward dialogue model, (2) a dual-paired variation with only paired data, and (3) the full DualEmp model with both paired and unpaired data. The evaluation is conducted on the EMPATHETICDIALOGUES dataset [24] with the split ratio of 8:1:1 for the training, validation, and test sets respectively.

For the offline evaluation metrics, the usual four ECS evaluation metrics accuracy (for emotion prediction), perplexity, BLEU, and distinct-1/distinct-2 are employed. In addition, embedding-based scores are used. The constructs empathy, relevance, and fluency are utilized for the user evaluation. Ablation study and human a/b testing results were also reported for the study. The evaluation results indicate the DualEmp model outperforms the baselines in all metrics in both offline and human evaluations. The ablation study results underline the utility of both the backward model and discrete latent variable. The dual-paired variant by itself outperforms all the baselines while the single-paired variant produces comparatively poor results. In the A/B testing, pairwise comparisons highlight that responses from DualEmp models are more preferred by humans than those from baselines.

## 5 GAPS AND OPPORTUNITIES

The contemporary approaches in ECS studies have helped immensely in improving the experience and perceptions of users during their interaction with ECS, albeit in an open-domain setting. In this section, we are highlighting the current gaps in ECS studies in the context of both open-domain and closed-domain settings.

**Aspect-level emotion identification.** This gap has been alluded to in a previous survey paper [35]. Many ECS studies have an emotion detection module before the empathetic response generation module. The detected emotion class influences the empathetic response generation. The emotion class is identified for each of the text messages of the users. In open-domain settings, this approach appears to be partially adequate. However, when ECS is deployed in a specific domain (e.g., customer service), the system needs to identify the entities (aspects) in the text for which the emotions are expressed. This research area is referred to as target-dependent emotion analysis or aspect-based emotion analysis (ABEA). Although the aspect-based sentiment analysis (ABSA) area has been researched extensively [78], [79], ABEA research has not received much attention.

**Empathetic behavior categories.** The empathetic responses generated by the current ECS approaches may not be suitable for all scenarios due to the nature of the datasets. For example, the training datasets in existing ECS studies have been prepared through (a) crowdsourcing, (b) annotating social media data (e.g., Reddit [25]), and (c) annotating publicly available relevant data (e.g. OpenSubtitles [27]). In these datasets, the empathetic responses of humans are either solely provided or augmented with ratings. Literature on empathetic mechanisms has shown that there are multiple types of empathetic behaviors, namely mirroring, affective matching, empathic concern, consolation, altruistic helping, and perspective-taking [22]. We posit that datasets should be annotated with empathetic behavior type labels so that models can generate more accurate empathetic responses, depending on the level of the empathy system.

**Empathy incorporation approach.** In current ECS studies, the empathetic responses are directly generated by the main response generation module (MRGM). Although the response generation models in this module are pre-trained with general-purpose dialogue datasets such as DailyDialog [99] and BookCorpus [101] in certain studies, the leading training datasets of conversations are annotated with emotion labels and/or empathy levels. Thus, the empathetic response is directly embedded in the primary response. Other empathy incorporation pathways can be studied. Pre-RGM and Post-RGM are two such pathways. In the Pre-RGM pathway, once the NLU model processes the user's message, the emotion class(es) of the user messages and the constituent entities along with the whole user message are passed over to an empathetic response generation model (ERGM). The ERGM generates an empathetic response in the form of a text snippet which is passed over to MRGM, which combines the empathetic snippet with the subject/intent specific response. In the case of the Post-RGM pathway, the inverse is attempted. First, the MRGM generates a response and then ERGM adds the empathetic part to the main response by rewriting the response. This empathy rewriting task is a recently proposed task that was proposed to improve the empathetic conversational abilities of counselors in mental health settings [130].

**Multimodal operationalization of conceptual empathy models and frameworks.** The conceptual empathy models and frameworks (covered in Section 3) provide a blueprint for different aspects that need to be considered for an effective, empathetic response generation process. It must be noted that the existing ECS studies have not fully operationalized all the features proposed in these models and frameworks. Multimodality is one such feature. The approaches proposed in the existing studies are based on text modality. While multimodal dataset availability is an ongoing challenge in this research topic, future studies should consider audio, and video inputs as full-scale empathy can be exhibited by ECS implementations primarily with multimodal inputs [32]. The second feature that current ECS studies have not considered is the multi-level empathy approach. As proposed in [32], this approach would help in implementing empathy at low, mid, and high levels. These levels are differentiated based on characteristics such as mimicry, affect matching, user mood, user personality, user likes/dislikes, user goals, and context.

**Integration with traditional CAI approaches for domain-specific use cases.** In the existing ECS studies, the main task is to generate empathetic responses in an open-domain setting (day-to-day usage). However, this approach is not entirely feasible in a domain-specific setting since user intent detection plays a crucial role in defining the responses. Few studies have employed commonsense KBs to extract concepts from the user's text message and deduce relationships between these concepts to generate more relevant empathetic responses [66]–[68]. This approach should work in a closed domain setting, in theory. Another approach is to blend different task types into a single model so that the model has an enhanced ability to respond to different user intents [131], [132]. In domain-specific CAI implementations, the NLU module is expected to detect the user intents and perform slot filling. These capabilities need to be incorporated when deploying domain-specific ECS.

## 6 CONCLUSION AND FUTURE WORK

This paper contributes to the topic of empathetic conversational dialogue systems. The state-of-the-art conceptual empathy models, frameworks, and empirical studies have been reviewed. There have been few albeit comprehensive attempts at creating theoretical empathy models based on social science theories related to empathy, such as the theory of mind, belief-desire-intention model, and perception-action-mechanism. Based on these theories, conceptual, functional models, and frameworks have been put forth to highlight the methods of extracting and processing meaningful features from multimodal input data of users, followed by affect recognition and empathetic response generation. We consider it noteworthy that the existing empirical ECS studies have not entirely operationalized all the ideas and features from the conceptual models and frameworks.

The empirical studies have been reviewed in operationalized concepts, datasets, request processing & response generation techniques, and evaluation approaches & metrics. Handpicked SOTA approaches in ECS studies were introduced to highlight the effectiveness of leveraging certain empathy-related concepts at the model design level. Under operationalized concepts, ECS studies have extracted multiple features from the user messages such as emotion, sentiment, emotion causes, persona, and entities & their relations using external knowledge bases. At the ECS messages level, studies have embedded additional features such as exemplars and emojis. To improve the communication between users and ECS agents, few studies have incorporated a feedback loop through active and adversarial learning.

The EMPATHETICDIALOGUES dataset is the seminal dataset used in most ECS studies. The active datasets are prepared through crowdsourcing or annotation of social media data (predominantly from Reddit). General-purpose dialogue datasets DailyDialog and BookCorpus have also been in certain studies for pre-training the response generation models. Transformer architectures are fairly prevalent in both request processing and response generation modules of ECS implementations. Mirroring the trend in the dialogue systems research landscape, ECS studies have incorporated the latest architectures such as T5 for improving the response generation accuracy in the models. Offline evaluation is the dominant evaluation approach employed in ECS studies. While user evaluation has been attempted in multiple studies, many evaluation metrics are quite limited and restricted to empathy, fluency, and relevance. There is scope to incorporate more in-depth user perception metrics, catering to other facets of the conversation. Among the carefully selected SOTA approaches, the significance of incorporating the emotion causes and external knowledge in the response generation process has been highlighted along with the affect matching mechanism. Since these concepts improve the empathetic abilities of the models, future ECS studies might benefit by incorporating all these three concepts in the model design. Taking the cue from these specific studies, future ECS studies should aim to design offline and human evaluation experiments with prominent baselines.

While the current ECS approaches are tailored for general-purpose day-to-day conversations, they are not directly suitable for a domain-specific context. We opine that target-dependent emotion identification should be performed for providing fine-grained responses. The empathetic responses could be more varied with concepts adopted from existing theoretical empathy models. Different empathy incorporation pathways are to have experimented with in the response generation module as the current integrated approach will not be suitable for all domain-specific scenarios.

As a part of future work, we will start by proposing a multidomain empathy framework and subsequently implement the framework for two domains:- customer service and mental health. In our framework, we will be including multi-level empathy systems definition and corresponding design features, empathy incorporation pathways, operational modules, and enhanced evaluation metrics. This framework will be applicable for implementation in most domain settings. In addition, we plan to design and implement target-dependent/aspect-level emotion identification and emotion authentication algorithms since accurate emotion understanding is a crucial step towards providing a comprehensive, empathetic response.

## ACKNOWLEDGMENT

This research is supported by the Agency for Science, Technology and Research (A*STAR) under its SERC Council Strategic Fund (C210415006). The authors are grateful for the helpful discussion with Raj Kumar Gupta and Ajay Vishwanath.

**Aravind Sesagiri Raamkumar** received the PhD degree in information studies from Nanyang Technological University, Singapore. He is currently a scientist with the Affective Computing Group, Social and Cognitive Computing Department, Institute of High Performance Computing, A*STAR, Singapore. His research interests include recommender systems, text mining, information retrieval, health informatics, social network analysis, scholarly metrics, deep learning, data mining, scholarly communication, social media and linked data.

**Yinping Yang** received the PhD degree in information systems from the National University of Singapore. She is currently a senior scientist and group manager with the Affective Computing Group, Social and Cognitive Computing Department, Institute of High Performance Computing, A*STAR, Singapore. She is the principal investigator with A*STAR's digital emotions and empathy machine programme and manages industry and public sector projects in emotion analytics and e-negotiation technologies. Her research interests include intelligent negotiation systems, emotion recognition, and strategic foresight.